\documentclass{article}

\usepackage{microtype}
\usepackage{graphicx}
\usepackage{subfigure}
\usepackage{booktabs} 

\usepackage{hyperref}

\usepackage[accepted]{icml2023}

\usepackage{amsmath}
\usepackage{amssymb}
\usepackage{mathtools}
\usepackage{amsthm}
\usepackage{diagbox}                                                                                       
\usepackage{graphicx}
\usepackage{bm}
\usepackage[capitalize,noabbrev]{cleveref}

\theoremstyle{plain}

\theoremstyle{definition}

\theoremstyle{remark}

\usepackage[textsize=tiny]{todonotes}
\usepackage{multicol}
\usepackage{multirow}

\icmltitlerunning{Triplet Knowledge Distillation}

\def\ie{\emph{i.e.}} 
 
\def\etc{\emph{etc}} 
 
\def\etal{\emph{et al.}}
\def\kdname{TriKD}
\begin{document}
\twocolumn[
\icmltitle{Triplet Knowledge Distillation}



\icmlsetsymbol{equal}{*}
\icmlsetsymbol{equal}{*}
\begin{icmlauthorlist}
\icmlauthor{Xijun Wang}{equal,yyy,ucas}
\icmlauthor{Dongyang Liu}{equal,yyy,ucas}
\icmlauthor{Meina Kan}{yyy,ucas}
\icmlauthor{Chunrui Han}{yyy,ucas}
\icmlauthor{Zhongqin Wu}{comp}
\icmlauthor{Shiguang Shan}{yyy,ucas,pengcheng}
\end{icmlauthorlist}

\icmlaffiliation{yyy}{Key Laboratory of
Intelligent Information Processing, Institute of Computing Technology (ICT),
Chinese Academy of Sciences (CAS)}
\icmlaffiliation{ucas}{University of
Chinese Academy of Sciences}
\icmlaffiliation{pengcheng}{Peng Cheng Laboratory}
\icmlaffiliation{comp}{Horizon Robotics}

\icmlcorrespondingauthor{Xijun Wang}{xijun.wang.cs@gmail.com}
\icmlcorrespondingauthor{Dongyang Liu}{dongyang.liu@vipl.ict.ac.cn}


\vskip 0.3in
]



\printAffiliationsAndNotice{\icmlEqualContribution} 

\begin{abstract}
In Knowledge Distillation, the teacher is generally much larger than the student, making the solution of the teacher likely to be difficult for the student to learn. To ease the mimicking difficulty, 
we introduce a triplet knowledge distillation mechanism named~\kdname{}. Besides teacher and student,~\kdname{} employs a third role called anchor model. Before distillation begins, the pre-trained anchor model delimits a 
subspace within the full solution space of the target problem. Solutions within the subspace are expected to be easy targets that the student could mimic well. Distillation then begins in an online manner, and the teacher is only allowed to express solutions within the aforementioned subspace. Surprisingly, 
benefiting from accurate but easy-to-mimic hints, the student can finally perform well. 
After the student is well trained, it can be used as the new anchor for new students, forming a curriculum 
learning strategy. Our experiments on image classification and face recognition with various models clearly demonstrate the effectiveness of our method. Furthermore, the proposed ~\kdname{} is also effective in dealing with the overfitting issue. Moreover, our theoretical analysis supports the rationality of our triplet distillation.
\end{abstract}
\section{Introduction}
Knowledge distillation (KD) generally optimizes a small student model by transferring knowledge from a large teacher model.
\begin{figure*}[h]
\normalsize
\centering
\includegraphics[width=1\linewidth]{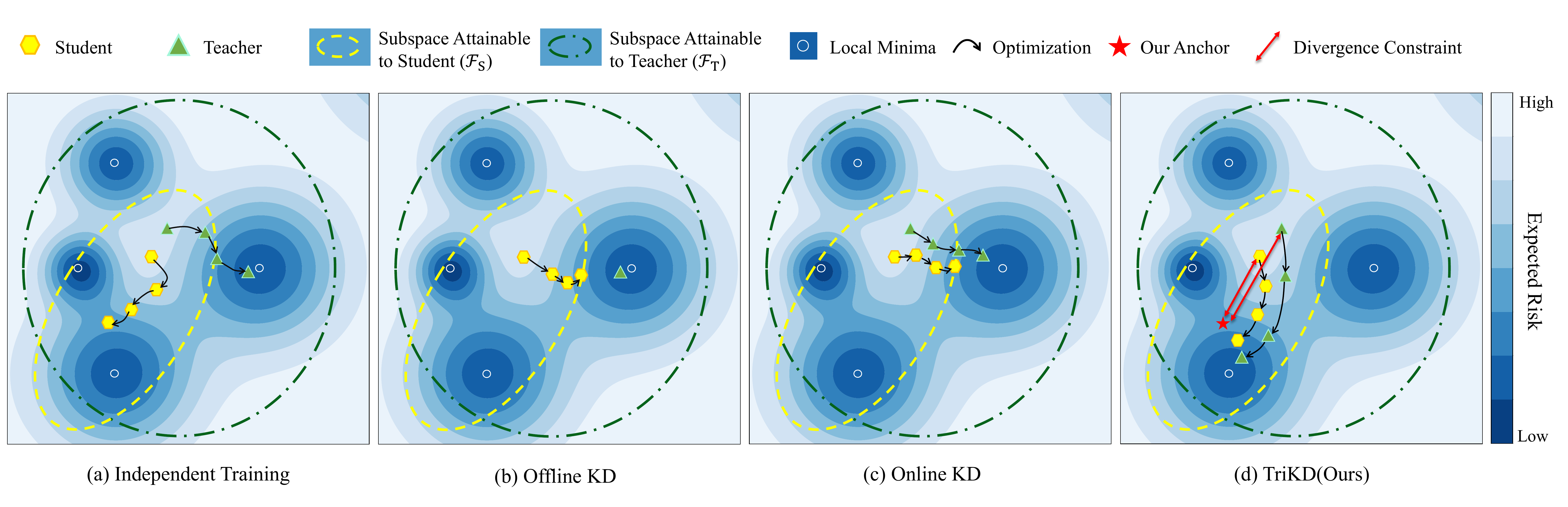}
\caption{An intuitive illustration of our motivation. The 2d plane represents the function space from input image to task-specific output. Every neural network with compatible input and output format corresponds to a certain point on the plane, and the color represents the expected risk, darker means lower risk. The small model is the target student and its performance is our major interest. As the large teacher model has stronger fitting ability than the student, the collection of functions it could attain, $\mathcal{F}_\mathrm{T}$, is also larger than $\mathcal{F}_\mathrm{S}$. (a) When trained independently, the teacher model may step towards local minima out of the scope that the student could well fit. (b)(c) For both online and offline distillation, the large model is likely to lie beyond the subspace attainable to student model. This makes the student, though performing better, still lie far away from the optima, leading to a sub-optimal solution. (d) In our TriKD, a pre-trained anchor model is used to pull both the teacher and student models within or near the subspace attainable to the student model, making the teacher easy to mimic. The mutual learning between teacher and student then makes the student learn a high-quality solution with better generalization. }
\label{fig:illustration}
\end{figure*}
While most existing works aim to make a student learn better from a given teacher, the training of the teacher itself usually follows the trivial way and is rarely investigated. However, without any intervention, large models suffer from high risk of coming into solutions that, while generalize well, are difficult for small models to mimic, which would unfavourably affect distillation. This argument is supported by recent work showing the optimization difficulty is a major barrier in knowledge distillation~\cite{reallywork}, and is also confirmed by evidence that larger teacher with higher accuracy counter-intuitively makes worse student~\cite{ESKD, StudentCustomized, teacherassistant}. An illustration is shown in Fig.\ref{fig:illustration}(a-c). Considering the function space from input image to target output, the subspace consisting of functions that the teacher could fit, $\mathcal{F}_\mathrm{T}$ (referred to as \textit{hypothesis space} in machine learning), is larger than that of the student, $\mathcal{F}_\mathrm{S}$, since the teacher has larger capacity. When the solution of the teacher is out of the subspace attainable to the student ($\mathcal{F}_\mathrm{S}$), the student would fail to mimic the teacher's solution well. 

Our proposed method, \kdname{}, is based on online knowledge distillation and inspired by the following motivation: \textit{could we make the teacher not only accurate, but also easy to mimic?} In this paper, we try to achieve this goal through providing both the online teacher and the student with a common anchor, which constrains the two models to learn to solve the target task in a small-model friendly approach. The pre-trained anchor model is of \textbf{equal} capacity comparing with the \textbf{student}, which ensures the function expressed by the anchor, $f_\mathrm{A}$, is within $\mathcal{F}_\mathrm{S}$ and easily mimickable to the student. By penalizing the function distances from the anchor to the student and especially to the teacher, the anchor pulls the search space of both the student and especially the teacher near $f_\mathrm{A}$. The teacher then has good chance to also lie within or close to $\mathcal{F}_\mathrm{S}$, leading to easy mimicking. Meanwhile, even being restricted to a small search space, we find that the large teacher could still reveal high-accuracy solutions thanks to its high capacity. Benefited from accurate but easy-to-mimic hints, the student can then mimic the teacher more faithfully and perform better after distillation. \textit{In short, the anchor model, teacher model, and student model formulate a novel triplet knowledge distillation mechanism.} An illustration is shown in Fig.\ref{fig:illustration}(d).

Since an appropriate anchor is not trivial to find, we develop a curriculum strategy: the trained student from one~\kdname{} generation is used as the anchor of the next generation, and a new pair of randomly initialized student and teacher join in. Generation by generation, the newly trained student becomes better and better, and its performance finally converges. Considering Fig.\ref{fig:illustration}(d), this process can be interpreted as gradually moving the anchor towards local minima.

Overall, \textit{our main contributions are as below}: 1).\noindent~We propose a novel triplet knowledge distillation mechanism named~\kdname{}.~\kdname{} makes distillation more efficient by making the teacher not only accurate by also easy to mimic. 2). To find a proper anchor model for~\kdname{}, we propose a curriculum strategy where student in one generation serves as the anchor of the next generation. 3). Our~\kdname{} achieves state-of-the-art performance on knowledge distillation, and also demonstrates better generalization in tackling the overfitting issue. 4). Theoretical analysis in a statistical perspective is given to analyze the rationality of triplet distillation.

\begin{figure*}[h]
\normalsize
\centering
\includegraphics[width=1\linewidth]{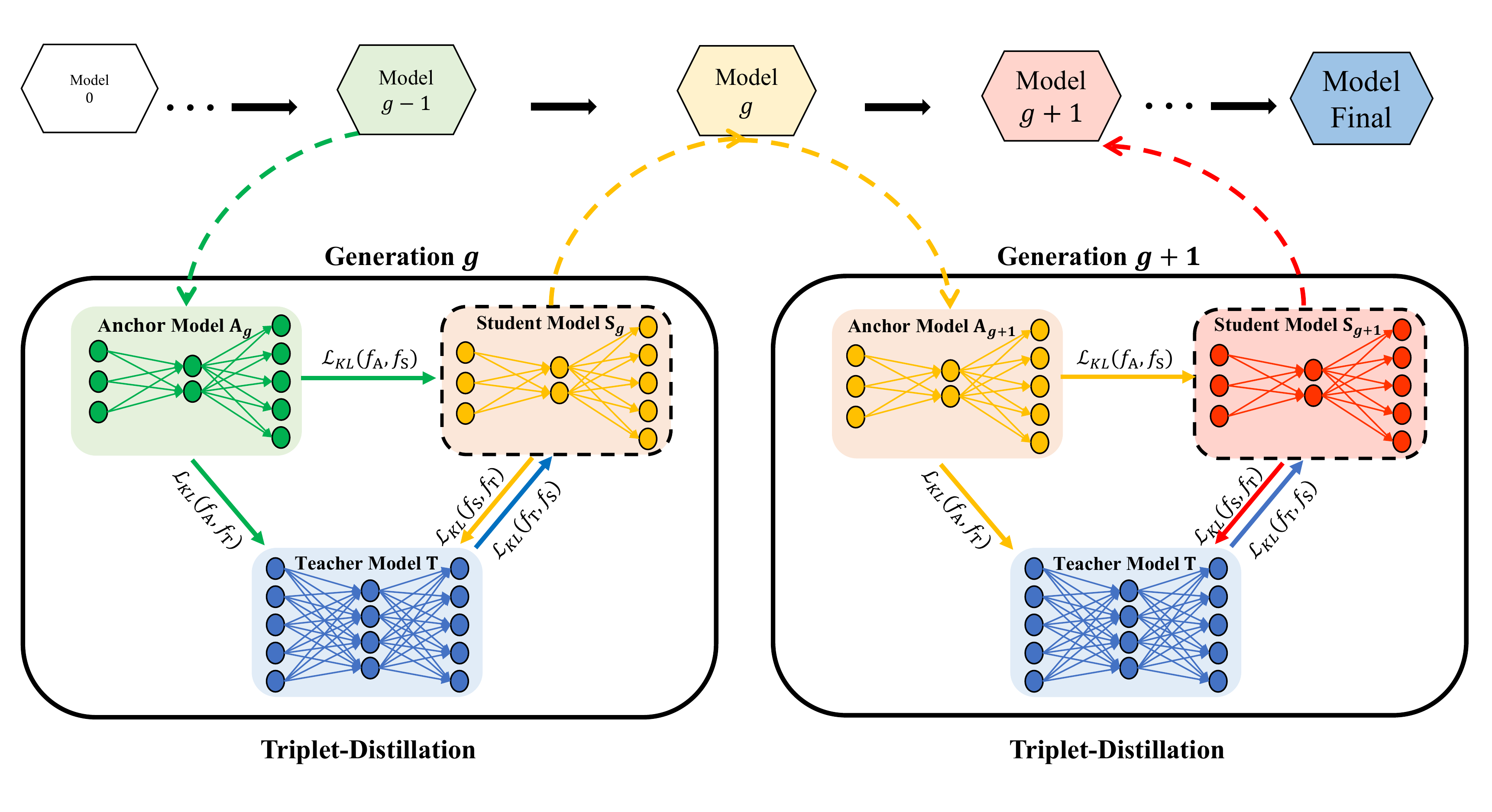}
\vspace{-1.2cm}
\caption{An overview of Triplet Knowledge Distillation. In the $g$th generation, a pre-trained anchor $\mathrm{A}_g$ supervises a pair of randomly initialized student $\mathrm{S}_g$ and teacher $\mathrm{T}_g$; the student and the teacher also learn mutually from each other. After the $g$th generation, the student $\mathrm{S}_g$ will become the new anchor $\mathrm{A}_{g+1}$ for the $(g+1)$th generation. Supervision from task label is omitted in the figure.}
\label{fig:pipeline}
\vspace{-0.3cm}
\end{figure*}

\section{Related work}
\subsection{Offline Knowledge Distillation}
Offline knowledge distillation makes the student learn from a \textbf{pre-trained} and \textbf{fixed} teacher. ~\citet{hinton2015distilling} propose mimicking the softened class distributions predicted by large teachers. Some studies~\cite{ding2019adaptive, wen2019preparing} then go a further step to explore the trade-off between the supervision of soft logits and hard task label, and others~\cite{tian2019contrastive, SSKD} propose to introduce auxiliary tasks to enrich the transferred knowledge. Instead of final outputs, many works exploit the intermediate features~\cite{romero2014fitnets, FT, jin2019knowledge, zagoruyko2016paying, review} as transferred knowledge. Self-distillation, pioneered by Born again~\cite{furlanello2018born}, makes the teacher share the same network architecture as the student, and continuously updates the student in an iterative manner. Our~\kdname{} is related to Born again as it also involves such iterative training, but we use it to obtain a more reliable anchor. 

\subsection{Online Knowledge Distillation} 
\normalsize
Online knowledge distillation makes multiple randomly-initialized models collaboratively learn from scratch. This line of research is especially significant for scenarios without available pre-trained teacher model. A monumental work is deep mutual learning (DML)~\cite{zhang2018deep}. During the training phase, DML uses a pool of randomly initialized models as the student pool, and each student is guided by the output of other peers as well as the task label. Based on DML, some works~\cite{AMLN, DCM} additionally take intermediate features into account, and others~\cite{KDCL, okddip} design different mimicking targets. Our~\kdname{} is also built upon DML as the teacher and the student are all randomly initialized and learn mutually from each other, but we additionally incorporate an anchor model to enhance distillation.

\subsection{'Larger Teacher, Worse Student'}
Intuitively, the performance of the student should increase when the teacher has larger capacity and higher performance. However, Cho~\etal{}~\cite{ESKD} identify that very large teacher actually makes the student deteriorate. This phenomenon has also been witnessed by following works~\cite{teacherassistant, StudentCustomized}, and has been attributed to the capacity mismatch between teacher and student. To overcome this problem, ESKD~\cite{ESKD} proposes an early-stopping strategy, and SCKD~\cite{StudentCustomized} automatically adjusts the distillation process through considering the gradient similarity between the teacher’s and the student's distillation loss. TAKD~\cite{teacherassistant} divides the distillation process into multiple stages, and introduces intermediate-sized models, called teacher assistant, to bridge the capacity gap between the original teacher and student. While TAKD~\cite{teacherassistant} treats mimicking difficulty as an inherent property of teacher model capacity,~\ie{}, larger teachers are inherently harder to mimic, we believe that a given large network  with fixed capacity should be able to fit both hard and
easy functions, and we could make a large teacher still easy to mimic by deliberately making the function it expresses easy. Detailed comparisons between TAKD and our TriKD are provided in~\ref{sec_takd_compare} in Appendix.

\section{Method}
\subsection{Triplet Distillation}
\label{method_trikd}
Our \kdname{} incorporates three models: online teacher~$\mathrm{T}$, student~$\mathrm{S}$, and anchor~$\mathrm{A}$. Among them, the anchor supervises both the teacher and student, and the student and the teacher learn mutually from each other. At the beginning of the distillation process, the anchor is already fully-trained on the target task, while the student and the teacher are randomly initialized. During distillation, the parameters of the anchor model keep fixed, while the parameters in the other two models are optimized, which is detailed below.

\subsubsection{\small{Guidance from anchor to teacher/student}} 
\label{method:trikd:anchor}
The anchor $\mathrm{A}$ is designed to constrain the student  $\mathrm{S}$ and the teacher $\mathrm{T}$ to learn to solve the target task in a student-friendly manner. For this purpose, we first ensure the function expressed by the anchor itself, $f_\mathrm{A}$, is easily attainable to the student. This is achieved by making the anchor model $\mathrm{A}$ of the same architecture and size as the student $\mathrm{S}$, and already trained on the target task. We then try to constrain the search space of both the teacher and the student to be near $f_{\mathrm{A}}$, which is realized through penalizing the KL-divergence from the anchor to the teacher/student:
\begin{equation}
\label{KL_AT}
\small
        \mathcal{L}_{KL}(f_\mathrm{A},f_\mathrm{T}) = \sum_{i=1}^{N}\tau^2 \mathbf{KL}\left(f_\mathrm{A}(x_i)||f_\mathrm{T}(x_i)\right),
\end{equation}
\begin{equation}
\label{KL_AS}
\small
        \mathcal{L}_{KL}(f_\mathrm{A},f_\mathrm{S}) = \sum_{i=1}^{N}\tau^2 \mathbf{KL}\left(f_\mathrm{A}(x_i)||f_\mathrm{S}(x_i)\right),
\end{equation}
where $x$ denotes training sample, $N$ is the number of training samples, $\tau$ represents temperature used to soften the output distributions. Specifically, 
\begin{equation}
\small
    f_{(\cdot)}(x) = \sigma(\frac{\bm{z}_{(\cdot)}(x)}{\tau}),
\end{equation}
where $\sigma$ denotes the softmax function, and $\bm{z}$ is logit scores output by the penultimate layer of the neural network. 
In this way, the teacher is prevented from solutions that are far from the anchor, and thus has good chance to lie within or close to $\mathcal{F}_\mathrm{S}$. It is then reasonable to expect that the function expressed by the teacher, $f_\mathrm{T}$, would be a relatively easy mimicking target to the student. We will show some experiment results supportive of this expectation in~\ref{similarity}, which demonstrate that the constraint from the anchor does make mimicking easier, as teacher-student behavior similarity becomes substantially higher.  

\subsubsection{Mutual distillation between teacher and student}

When not considering the anchor $\mathrm{A}$, the rest part of~\kdname{} follows the standard online knowledge distillation method DML~\cite{zhang2018deep}. Specifically, the student and the online teacher not only learn from the hard labels, but also mutually draw lessons from the training experiences of each other. From the student perspective, the loss regarding hard label is the standard cross-entropy loss $\mathcal{L}_{ce}(f_\mathrm{S})$, defined as:
\begin{equation}
\small
    \mathcal{L}_{ce}(f_\mathrm{S}) = -\sum_{i=1}^{N}\sum_{k=1}^{K} y_i^k log(f_\mathrm{S}^k(x_i)),
\end{equation}
$K$ is the number of classes, $y$ is hard classification label. Furthermore, the student also learns from the teacher:
\begin{equation}
\small
    \mathcal{L}_{KL}(f_\mathrm{T},f_\mathrm{S}) = \sum_{i=1}^{N}\tau^2 \mathbf{KL}\left(f_\mathrm{T}(x_i)||f_\mathrm{S}(x_i)\right).
\end{equation}
Combining with the constraint from anchor, the complete loss function for the student is:
\begin{equation}
\label{loss_stu}
\small
    ~\mathcal{L}_{\mathrm{S}} = w_1 \mathcal{L}_{ce}(f_{\mathrm{S}}) + w_2 \mathcal{L}_{KL}(f_\mathrm{T},f_\mathrm{S}) + w_3 \mathcal{L}_{KL}(f_\mathrm{A},f_\mathrm{S}).
\end{equation}
Similarly, the loss function for the teacher is in the symmetric form:
\begin{equation}
\label{loss_tea}
\small
    ~\mathcal{L}_{\mathrm{T}} = w_4 \mathcal{L}_{ce}(f_{\mathrm{T}}) + w_5\mathcal{L}_{KL}(f_\mathrm{S},f_\mathrm{T}) + w_6 \mathcal{L}_{KL}(f_\mathrm{A},f_\mathrm{T}),
\end{equation}
where $w$ is the weight of each loss. For $\mathcal{L}_{ce}$, $\tau$ is fixed to $1$, whereas for $\mathcal{L}_{KL}$, $\tau$ is a hyper-parameter to tune.

Our~\kdname{} is based on online knowledge distillation, and uses an additional anchor to make the teacher easy to mimic by constraining the search space. On the other hand, we hope the teacher, with large capacity and correspondingly strong learning ability, could still find a low-expected-risk solution to accurately guide the student, even though its search space is constrained by the anchor. Note that here exists a potential risk that if the constraint from the anchor is too strong ($w_3$ and $w_6$ are too large), the performance of the teacher may be upper-bounded by the anchor, thus leading to easy but inaccurate teacher solutions. However, experiments in~\ref{similarity} and~\ref{teacher-performance} show that with proper hyper-parameters, the teacher can be both easy (\ref{similarity}) and accurate (\ref{teacher-performance}) simultaneously. This means that low mimicking difficulty of the teacher could be attained even when the constraint from anchor is relatively mild, and the constraint would not barrier the accuracy of the teacher until its grows much stronger. There is thus a range of constraint strength where the merits of both low-mimicking-difficulty and low-expected-risk teacher could be simultaneously enjoyed.  With the aforementioned merits, the student could benefit substantially more from~\kdname{} than existing distillation methods, and finally become more accurate than existing models.

\subsection{Curriculum learning for Proper Anchor}
\label{iterationsec}
Intuitively, the selection of anchor model affects the performance of~\kdname{}, and it is thus of great significance to find a proper anchor. However, such an appropriate anchor is not trivial to find. We therefore propose a curriculum strategy to achieve this goal.

The curriculum process is composed of a sequence of \textbf{generations}, each of which is a triplet distillation process as described in~\ref{method_trikd}. In curriculum learning, the student of the $g$th generation will become the anchor of the $(g+1)$th generation, denoted as: 
\begin{equation}
    \mathrm{A}_{g+1} = \mathrm{S}^*_g,
\end{equation}
where $\mathrm{S}^*_g$ is the student trained in the $g$th generation. The student and the teacher are randomly re-initialized at the beginning of each generation. We empirically find that the performance of the student tend to raise within the first several generations; it then converges and more generations would not make further improvement. We can then take the student with converged performance as the final model, which is generally with better performance. Fig.\ref{fig:pipeline} shows the whole pipeline of the proposed method.

For the first generation, as there is no available last-generation student to serve as the anchor, we simply pre-train the anchor model with only online distillation between it and the teacher. We also try to use a trivial one only trained with label, and find it achieves comparable performance but with slower convergence. Therefore, in this paper we use the student trained with vanilla online distillation as the anchor for generation 1, and we refer to the vanilla online distillation process itself as generation 0. 

\subsection{Theoretical Analysis}
\label{sec:intro-theory}
We explain why~\kdname{} could improve knowledge distillation in a formal context of the risk minimization decomposition. Lopez-Paz~\etal{}~\cite{unifying_vapnik} decomposed the excess risk of the student trained only with hard label as follows:
\begin{equation}
\small
    R(f_\mathrm{S}) - R(f_{\mathrm{R}}) \leq O\left(\frac{|\mathcal{F}_\mathrm{S}|_C}{\sqrt{n}}\right) + \epsilon_1,
\end{equation}
where $R(\cdot)$ denotes expected risk, $f_\mathrm{S}$ is the student function in function class $\mathcal{F}_\mathrm{S}$, $f_{\mathrm{R}}$ is the real (target) function. The $O(\cdot)$ term is the estimation error, and $\epsilon$ term is approximation error. $|\cdot|_C$ is some appropriate capacity measurement of function class. For distillation, the teacher learns from the target function, leading to the following excess risk:
\begin{equation}
\small
    R(f_\mathrm{T}) - R(f_{\mathrm{R}}) \leq O\left(\frac{|\mathcal{F}_\mathrm{T}|_C}{n^{\alpha}}\right) + \epsilon_2,
\end{equation}
and the student learns from the teacher,  leading to the following excess risk:
\begin{equation}
\small
    R(f_\mathrm{S}) - R(f_{\mathrm{T}}) \leq O\left(\frac{|\mathcal{F}_\mathrm{S}|_C}{n^{\beta}}\right) + \epsilon_3,
\end{equation}
where $\alpha, \beta$ range between $[\frac{1}{2}, 1]$, higher value means easier problem and faster learning. As analyzed in~\cite{unifying_vapnik}, the effectiveness of vanilla knowledge distillation is theoretically ensured by the following inequality:
\begin{equation}
\label{kdcondition}
\small
     O\left(\frac{|\mathcal{F}_\mathrm{T}|_C}{n^{\alpha}}\right) +  O\left(\frac{|\mathcal{F}_\mathrm{S}|_C}{n^{\beta}}\right) + \epsilon_2 + \epsilon_3 \leq O\left(\frac{|\mathcal{F}_\mathrm{S}|_C}{\sqrt{n}}\right) + \epsilon_1.
\end{equation}
Furthermore, if the left side of Eq. (\ref{kdcondition}) decreases, the excess risk of the student becomes lower, meaning better performance. Next, we show that introducing the anchor model $\mathrm{A}$ lowers the left side of Eq. (\ref{kdcondition}).

Considering vanilla online knowledge distillation, its loss function is:
\begin{equation}
\small
\begin{aligned}
    \mathcal{L}_{online}=&w_1 \mathcal{L}_{ce}(f_{\mathrm{S}}) + w_2 \mathcal{L}_{KL}(f_\mathrm{T},f_\mathrm{S}) 
         \\ & + w_4 \mathcal{L}_{ce}(f_{\mathrm{T}}) + w_5 \mathcal{L}_{KL}(f_\mathrm{S},f_\mathrm{T}).
\end{aligned}
\end{equation}
\kdname{} can be equivalently recognized as minimizing $\mathcal{L}_{online}$, but with additional inequality constraints coming from the anchor:
\begin{equation}
\small
\begin{aligned}
\label{new_problem}
        \underset{f_\mathrm{S}, f_\mathrm{T}}{min} \quad & \mathcal{L}_{online},\\
        s.t. \quad & \mathcal{L}_{KL}(f_\mathrm{A},f_\mathrm{S}) < \delta, \\
        \quad & \mathcal{L}_{KL}(f_\mathrm{A},f_\mathrm{T}) < \delta,
\end{aligned}
\end{equation}
where $\mathcal{L}_{KL}$ serves as a function distance metric to constrain the search space of the teacher and the student; $\delta$ is the distance threshold. Rather than directly solving~Eq. (\ref{new_problem}), we can instead add penalty terms to the loss function to substitute the hard constraints, making the optimization much easier. We then get Eq. (\ref{loss_stu}) and Eq. (\ref{loss_tea}), which we actually optimize in practice. Considering~Eq. (\ref{new_problem}), it means conducting the vanilla online distillation, but with constraints that shrink the search space of teacher $\mathrm T$ from the entire $\mathcal{F}_{\mathrm{T}}$ to its subset $\mathcal{F}_{\mathrm{T}}^{'}$:
\begin{equation}
\small
        \mathcal{F}_{\mathrm{T}}^{'} = \{f | f \in \mathcal{F}_{\mathrm{T}}, \mathcal{L}_{KL}(f_\mathrm{A},f_\mathrm{T}) < \delta\},
\end{equation}
and similarly shrink the search space of student $\mathrm{S}$ from $\mathcal{F}_{\mathrm{S}}$ to its subset $\mathcal{F}_{\mathrm{S}}^{'}$:
\begin{equation}
\small
        \mathcal{F}_{\mathrm{S}}^{'} = \{f | f \in \mathcal{F}_{\mathrm{S}}, \mathcal{L}_{KL}(f_\mathrm{A},f_\mathrm{S}) < \delta\}.
\end{equation}
The student and especially the teacher are then asked to find a solution within the shrinked search space $\mathcal{F}_{\mathrm{S}}^{'}$ and $\mathcal{F}_{\mathrm{T}}^{'}$. Following the left side of Eq. (\ref{kdcondition}), the risk bound for our proposed~\kdname{} is:
\begin{equation}
\label{trikdbound}
\small
     O\Big(\frac{|\mathcal{F}_\mathrm{T}^{'}|_C}{n^{\alpha^{'}}}\Big) +  O\Big(\frac{|\mathcal{F}_\mathrm{S}^{'}|_C}{n^{\beta^{'}}}\Big) + \epsilon_2^{'} + \epsilon_3^{'}.
\end{equation}
First, as $\mathcal{F}_{\mathrm{S}}^{'}$, $\mathcal{F}_{\mathrm{T}}^{'}$ are subsets of  $\mathcal{F}_{\mathrm{S}}$, $\mathcal{F}_{\mathrm{T}}$, we have $|\mathcal{F}_\mathrm{S}^{'}|_C \leq |\mathcal{F}_\mathrm{S}|_C$,  $|\mathcal{F}_\mathrm{S}^{'}|_C \leq |\mathcal{F}_\mathrm{S}|_C$. Next, recall that~\kdname{} is built upon two empirically-validated expectations: 1) the teacher would be easy to mimic if its search space is near $f_\mathrm{A}$ (\ie{} it is taken from $\mathcal{F}_{\mathrm{T}}^{'}$ rather than $\mathcal{F}_{\mathrm{T}}$), and 2) even the search space is constrained to $\mathcal{F}_{\mathrm{T}}^{'}$, the teacher could still find a low-expected-risk solution therein to provide accurate enough guidance. The first one implies that $\beta^{'}>\beta$,~\ie{} the mimicking from student to teacher is easier in our case. The second one implies that 
\begin{equation}
\label{eq:hypo2}
\small
    O\Big(\frac{|\mathcal{F}_\mathrm{T}^{'}|_C}{n^{\alpha^{'}}}\Big) + \epsilon_2^{'} \approx O\Big(\frac{|\mathcal{F}_\mathrm{T}|_C}{n^{\alpha}}\Big) + \epsilon_2,
\end{equation}
indicating the teacher would present similar expected risk either with or without anchor. Now we have analyzed all the involved variables except the $\epsilon_3$ term, and they all support that the bound in Eq. (\ref{trikdbound}) is lower than the left side of Eq. (\ref{kdcondition}). Finally, considering $\epsilon_3$ term, it signifies the approximation error from the student search space $\mathcal{F}_\mathrm{S}$ to the teacher function $f_\mathrm{T} \in \mathcal{F}_\mathrm{T}$:
\begin{equation}
\small
    \epsilon_3 = \Big(\mathop{\inf}_{f \in \mathcal{F}_{\mathrm{S}}}R(f)\Big) - R(f_{\mathrm{T}}).
\end{equation}
According to Eq. (\ref{eq:hypo2}), the difference in the $R(f_{\mathrm{T}})$ term will be minor between~\kdname{} and standard distillation; For the infimum term, in~\kdname{} $\mathcal{F}_{\mathrm{S}}^{'}$ replaces $\mathcal{F}_{\mathrm{S}}$, and since $\mathcal{F}_{\mathrm{S}}^{'}$ is a subset of $\mathcal{F}_{\mathrm{S}}$, its infimum should be higher, making $\epsilon_3^{'} \geq \epsilon_3$. However, it is unclear how large the difference is because the infimum on $\mathcal{F}_{\mathrm{S}}^{'}$ could still be very low. More importantly, the impact of the $\epsilon_3$ term to the total distillation process is limited, because the expected risk of real models in practice are far from the best one they could theoretically attain. Therefore, the influence of the $\epsilon_3$ term should be dwarfed by that of the other terms. \textit{Combining all the aforementioned changes together, the bound in Eq. (\ref{trikdbound}) is lower than the left side of Eq. (\ref{kdcondition}), signifying better distillation}. 


%
\begin{table*}[]
\caption{Compare the top-1 accuracy (\%) of different KD methods on CIFAR100. \textbf{Bold} and \underline{underline} denote the best and the second best results, respectively. For methods from KD to CRD, we quote the results in Tian~\etal{}~\cite{tian2019contrastive}. For Review to DKD, we show the results reported by their original authors. For DML, we report our reimplemented results. "$(\cdot)$" means the result was not reported by the authors and we re-run their provided codes. Note that DML and TriKD do not involve pre-trained teacher model.}
\label{cifar_main}
\small
\centering
\begin{tabular}{cccccccc}
\toprule
\multicolumn{1}{c}{\textbf{Teacher}}      & \textbf{wrn-40-2} & \textbf{wrn-40-2} & \textbf{resnet56} & \textbf{resnet110}  & \textbf{resnet110} & \textbf{resnet32x4} & \textbf{vgg13}             \\
\multicolumn{1}{c}{\textbf{Student}}      & \textbf{wrn-16-2} & \textbf{wrn-40-1} & \textbf{resnet20} & \textbf{resnet20}   & \textbf{resnet32}  & \textbf{resnet8x4}  & \textbf{vgg8}              \\ \midrule
\multicolumn{1}{c}{Teacher}               & 75.61             & 75.61             & 72.34             & 74.31               & 74.31              & 79.42               & 74.64                      \\
\multicolumn{1}{c}{Student}               & 73.26             & 71.98             & 69.06             & 69.06               & 71.14              & 72.50               & 70.36                      \\ \midrule
KD\cite{hinton2015distilling}             & 74.92             & 73.54             & 70.66             & 70.67               & 73.08              & 73.33               & 72.98                      \\
FitNet\cite{romero2014fitnets}            & 73.58             & 72.24             & 69.21             & 68.99               & 71.06              & 73.50               & 71.02                      \\
AT\cite{zagoruyko2016paying}              & 74.08             & 72.77             & 70.55             & 70.22               & 72.31              & 73.44               & 71.43                      \\
DML\cite{zhang2018deep}                   & 75.41             & 74.73             & 71.22             & 71.47               & 73.52              & 75.36               & 74.58                     \\
VID\cite{vid}                             & 74.11             & 73.30             & 70.38             & 70.16               & 72.61              & 73.09               & 71.23                      \\
CRD\cite{tian2019contrastive}             & 75.64             & 74.38             & 71.63             & 71.56               & 73.75              & 75.46               & 74.29                      \\
Review\cite{review}                       & 76.12             & \underline{75.09} & 71.89             & \underline{(71.86)} & 73.89              & 75.63               & \underline{74.84}          \\
DKD\cite{DKD}                             & \underline{76.24} & 74.81             & \underline{71.97} & (71.66)             & \underline{74.11}  & \underline{76.32}   & 74.68       \\
\midrule
TriKD(Ours)                               & \textbf{76.94}    & \textbf{75.96}    & \textbf{72.34}    & \textbf{72.55}      & \textbf{74.31}     & \textbf{76.82}      & \textbf{75.35}             \\
\bottomrule
\end{tabular}
\end{table*}

\begin{table*}[]
\normalsize
\caption{Compare different KD methods on ImageNet. \textbf{Bold} and \underline{underline} denote the best and the second best results, respectively. The results of Review of and DKD are from their original paper. Results of other existing methods are quoted from~\citet{tian2019contrastive}}
\label{imagenet_main}
\small
\centering
\setlength{\tabcolsep}{1mm}{
\begin{tabular}{c|cc|ccccccc|c}
    \toprule
    {\diagbox[innerwidth=2cm]{Error(\%)}{Methods}} & Teacher & Student & KD    & AT  &OFD    & CRD   & Review & DKD & DML & TriKD(Ours) \\ \midrule
    Top-1 & 73.31   & 69.75   & 70.66 & 70.69 & 70.81  & 71.17 & 71.61              & \underline{71.70} &   71.18     & \textbf{71.88}    \\ \midrule
    Top-5 & 91.42   & 89.07   & 89.88 & 90.01 & 89.98  & 90.13 & \underline{90.51}  & 90.41             &   90.05     & \textbf{90.70}    \\ \bottomrule
\end{tabular}
}
\end{table*}
\section{Experiments}
In this section, we empirically validate our proposed methods from five aspects. In~\ref{cifarresultsec} we compare~\kdname{} with state-of-the-art knowledge distillation methods on image classification to show the general effectiveness of the proposed method. In~\ref{growexpsec}, we validate the proposed method on the fine-grained problem of face recognition, with a special focus on the method's performance when confronting overfitting. In~\ref{similarity} and~\ref{teacher-performance}, we justify the rationality of our motivation. Specifically, in~\ref{similarity}, we show~\kdname{} makes the teacher an easier mimicking target from perspective of teacher-student behavior similarity; in~\ref{teacher-performance} we show the performance of the teacher is not limited by the small volume of $\mathcal{F}_\mathrm{T}^{'}$. In~\ref{ablation_study}, we conduct ablation studies to dissect the effect of each involved component. Detailed descriptions of experiment settings, as well as additional experiments and ablations, are provided in the Appendix.
\subsection{Knowledge Distillation on Image Classification}
\label{cifarresultsec}
We compare~\kdname{} with state-of-the-art knowledge distillation methods on two widely-used image classification benchmarks: CIFAR100~\cite{{krizhevsky2009learning}} and ImageNet~\cite{deng2009imagenet}. Given a pair of model architectures including one large and one small, we choose the small model as the anchor and as the student, and choose the big model as the teacher. 

\textbf{CIFAR100}~\cite{krizhevsky2009learning}: results are shown in Table~\ref{cifar_main}.~\kdname{} averagely raises the student's performance by 3.84\% comparing with the non-distillation baseline, and performs significantly better than vanilla KD~\cite{hinton2015distilling}, with an average improvement by 2.16\%.~\kdname{} also outperforms state-of-the-art methods on all teacher-student pairs. Note that~\kdname{} only uses the logits for knowledge transfer, but achieves better performance than those involving more complex information like intermediate feature map~\cite{review, romero2014fitnets, vid}, attention map~\cite{zagoruyko2016paying}, instance similarity~\cite{tian2019contrastive},~\etc{}

\textbf{ImageNet}~\cite{deng2009imagenet}: to validate the efficacy of our method on large-scale datasets, we also compare~\kdname{} with other methods on ImageNet. As shown in Table~\ref{imagenet_main},~\kdname{} also outperforms other methods, showing that the proposed triplet distillation mechanism could steadily produce high-quality models regardless of dataset volume.

\subsection{Knowledge Distillation on Face Recognition}
\label{growexpsec}
We validate our proposed~\kdname{} framework on the fine-grained problem of face recognition, with MobileFaceNet~\cite{chen2018mobilefacenets} as the main architecture. We use CASIA-WebFace~\cite{yi2014learning} for training, and MegaFace~\cite{kemelmacher2016megaface} for testing. Rank-1 face identification rate is reported. 

Unlike CIFAR100 and ImageNet, where the performance generally raises as the capacity of the model increases (at least within the scope of our interest), training with the CASIA-WebFace dataset is frequently bothered with the overfitting problem since each person has only about 50 images, which is much smaller than that on general image dataset. Intuitively, the constraint from the anchor prevents the teacher from expressing overly complicated functions. Therefore, we naturally wonder if~\kdname{} could help alleviate the overfitting issue. Consequently, for experiments on face recognition, we especially care about the relationship between student capacity and  performance. We fix the model size of teacher, but adjust the model size of student to investigate the relationship. For sake of convenience, in each generation we make the anchor model $\mathrm{A}$ slightly smaller than the student model $\mathrm{S}$, so that with training only one time we can obtain a serious of output models with increasing size. In all experiments unless otherwise specified, the student model starts with width 0.5X of MobileFaceNet and each generation uniformly increases the width of the network by 0.125 times of the MobileFaceNet size. The teacher model is 2.0X of MobileFaceNet  in all generations.

\begin{figure}[t]
\normalsize
\centering
\includegraphics[width=1\linewidth]{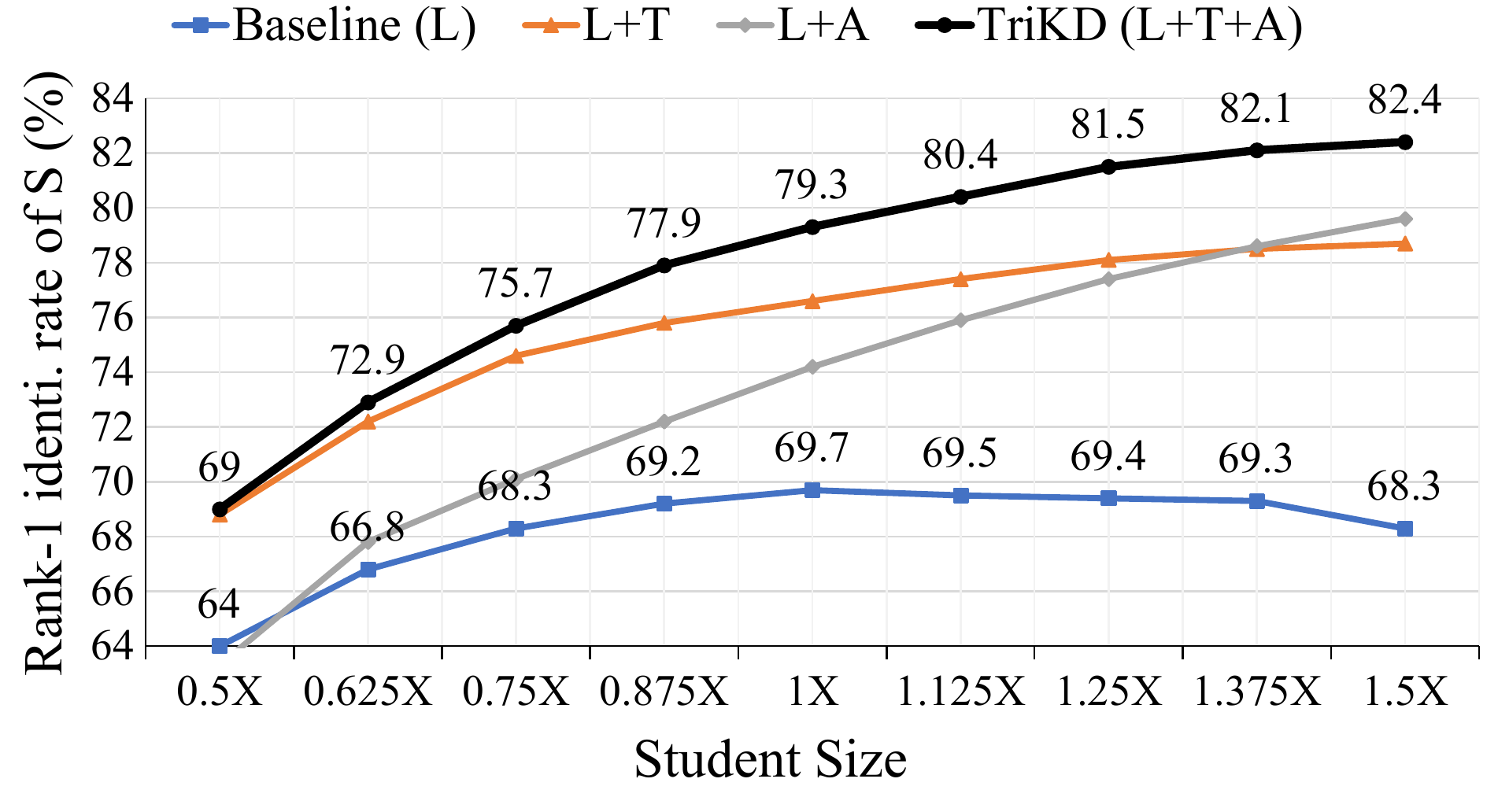}
\caption{Evaluate~\kdname{} with different student size on Megaface in terms of rank-1 face identification rate (\%). The baseline is trained with hard label only. Besides the baseline and our~\kdname{}, we also conduct ablative studies ($\mathrm{L}+\mathrm{T}$ and $\mathrm{L}+\mathrm{A}$) to reveal the effect of anchor $\mathrm{A}$ and $\mathrm{T}$, respectively. }
\label{figure_roles}
\end{figure}
\begin{table}[t]
  \small
  \caption{Comparison with existing methods on MegaFace in terms of rank-1 face identification rate (\%). Training set: CASIA-WebFace. Backbone: MobileFaceNet.}
  \label{result_webface}
  \centering
  \resizebox{\linewidth}{!}{
  \setlength{\tabcolsep}{1mm}{
  \begin{tabular}{c|cccc|c}
    \toprule
     {\diagbox[innerwidth=2cm]{Dataset}{Methods}}&baseline&KD&DML&BYOT&\kdname{} (Ours) \\
     \midrule
     50k&35.24&40.48&46.76&44.26&55.95 \\
    \hline
    150k&64.00&71.80&74.10&72.80&79.30   \\
    \hline
    490k&81.50&83.00&83.60&81.50&84.50 \\
    \bottomrule
  \end{tabular}}
  }
\end{table}

We first investigate the performance of the student w.r.t. its capacity. The 150k CASIA-WebFace subset is used for this experiment. The results are shown in Fig.\ref{figure_roles}. The \textbf{Baseline(L)} with only task label loss performs poorly, and starts in an underfitting state and then grows to an overfitting state. In contrast, our \textbf{\kdname{}} not only performs better than the baseline by a large margin in terms of all model sizes (even up to 10\% in G5, MobileFaceNet 1.125X), but also overcomes the overfitting issue, making the performance consistently raise as model capacity grows. Ablative results are also shown in Fig.\ref{figure_roles}, indicating both the teacher and the anchor are indispensable. We defer detailed analysis of this ablation study to Sec.\ref{ablation_study}. 

We further compare \kdname{} with the existing methods including KD~\cite{hinton2015distilling}, DML~\cite{zhang2018deep}, and BYOT~\cite{zhang2019your}. The 50k, 150k subsets and the full set with 490k images of CASIA-WebFace are used for training. The experimental results are shown in Table~\ref{result_webface}. As can be seen, our~\kdname{} achieves better accuracy. Importantly, the advantage of~\kdname{} is more significant with fewer training data: on the 490k training set, \kdname{} improves over the baseline by 3\%, and outperforms DML by 0.9\%; on the 50k training set, our TriDMG achieves larger improvement by 20.7\% comparing with the baseline, and by 9.19\% comparing with DML. The advantage in small-data tasks again indicates that~\kdname{} could help alleviate the overfitting problem.

\subsection{Teacher-Student Behavior Similarity}
\label{similarity}
We introduce the anchor $\mathrm{A}$ in hopes that it could lower the difficulty for the student to mimic the teacher. If it does work as expected, we should see an increase in teacher-student behavior similarity because the student would mimic the teacher more faithfully. Here we conduct experiments to validate this phenomenon. 

We show the KL-divergence between outputs of the student and the teacher trained on CIFAR100. For in-domain data, we report the results on CIFAR100. For out-of-domain data, where the student is more likely to act differently from the teacher, we report the results on SVHN~\cite{SVHN} and STL10~\cite{stl10}. Table~\ref{tab:similarity} shows the results. Compared with offline knowledge distillation, online distillation has a huge advantage in increasing teacher-student behavior similarity.
On the other hand, our~\kdname{} steadily shows great improvement upon online distillation, showing that the anchor does make the mimicking easier. The increase in teacher-student behavior similarity shows that the anchor model successfully drives the large teacher into easy-to-mimic solutions, supporting the expectation in~\ref{method:trikd:anchor}. 
\begin{table}[t]
\caption{Teacher-student behavior similarity on CIFAR100. Format: KL-divergence on training set/ KL-divergence on test set. Lower KL-divergence signifies stronger behavior similarity.}
\label{tab:similarity}
\small
\centering
\resizebox{\linewidth}{!}{
\setlength{\tabcolsep}{1mm}{
\begin{tabular}{c|ccccc}
\toprule
\multicolumn{1}{c|}{\multirow{2}*{Methods}}  &   wrn-40-2 & wrn-40-2  & resnet56  & resnet32x4 \\
      & wrn-16-2 & wrn-40-1 & resnet20 &  resnet8x4  \\ \midrule
Offline KD & 0.315/0.721 & 0.335/0.934 & 0.485/0.710 & 0.339/0.799  \\
Online KD & 0.088/0.228 & 0.094/0.233 & 0.133/0.205 & 0.075/0.247 \\
\kdname{}(Ours)   & 0.062/0.161 & 0.070/0.169 & 0.086/0.146 & 0.055/0.173 \\
\bottomrule
\end{tabular}}}
\end{table}
\begin{table}[t]
\caption{Teacher-student behavior similarity on SVHN and STL10. Format: KL-divergence on SVHN/ KL-divergence on STL10. Both on the test set. Lower KL-divergence signifies stronger teacher-student behavior similarity.}
\label{tab:similarity2}
\small
\centering
\resizebox{\linewidth}{!}{
\setlength{\tabcolsep}{1mm}{
\begin{tabular}{c|ccccc}
\toprule
\multicolumn{1}{c|}{\multirow{2}*{Methods}}  &   wrn-40-2 & wrn-40-2  & resnet56 & resnet32x4 \\
      & wrn-16-2 & wrn-40-1 & resnet20   & resnet8x4  \\ \midrule
Offline KD & 2.601/2.498 & 3.644/3.416 & 2.610/2.478 & 2.248/2.211\\
Online KD & 0.998/0.942 & 1.439/1.301 & 0.959/0.888 & 1.000/0.940 \\
\kdname{}(Ours)   & 0.761/0.711 & 1.096/0.987 & 0.673/0.625 & 0.726/0.680 \\
\bottomrule
\end{tabular}}}
\end{table}

\subsection{Performance of Teacher after~\kdname{}}
\label{teacher-performance}
\begin{table}[t]
\caption{\textbf{Teacher} Top-1 accuracy on CIFAR-100. Vanilla means trained with task labels only. Online means online distillation.}
\label{table:teacher_acc}
\small
\centering
\resizebox{\linewidth}{!}{
\setlength{\tabcolsep}{1mm}{
\begin{tabular}{c|ccccccc}
\toprule
{Teacher}      & {wrn-40-2} & {wrn-40-2} & {resnet56} & {resnet32x4} & {vgg13} \\
Student      & {wrn-16-2} & {wrn-40-1} & {resnet20} & {resnet8x4}  & {vgg8}  \\ \midrule
Vanilla               & 75.61             & 75.61             & 72.34              & 79.42               & 74.64          \\
Online KD & 77.74	& 78.05 & 74.00 & 80.28 & 75.91 \\
\kdname{}(Ours) & 79.01 & 78.70 & 75.12 & 80.05 & 76.09 \\
\bottomrule
\end{tabular}}}
\vspace{-3mm}
\end{table}
In~\kdname{}, the search space of the teacher is constrained by the anchor, and the teacher is expected to find a high-quality solution within the designated search space. This implies our expectation that the anchor would not barrier the teacher in chasing good solutions. Here we investigate the performance of teacher after~\kdname{} to check if the expectation holds. The results are shown in Table~\ref{table:teacher_acc}. The teacher actually outperforms its trivially-trained baseline, and also performs better than online distillation in most cases. The result indicates that the teacher is not encumbered by the constraint from anchor, and thus with~\kdname{}, we can simultaneously enjoy the merits of an easy-to-mimic and accurate teacher model. Note that existing works have already shown that online knowledge distillation would make both the large model (teacher) and the small model (student) improve~\cite{zhang2018deep}. However, it is also shown in~\cite{tian2019contrastive} that after switching from offline distillation to online distillation, the performance gain of the teacher could hardly trigger performance gain of the student. Our~\kdname{}, in contrast, makes the accurate teacher model also easy to mimic, and thus the student could benefit more from distillation.

\subsection{Ablation study}
\label{ablation_study}
The proposed triplet distillation consists of three roles, \ie{} the teacher $\mathrm{T}$, and target student $\mathrm{S}$ and the Anchor $\mathrm{A}$. From the student perspective, it is supervised by $\mathrm{T}$, $\mathrm{A}$ and task label $\mathrm{L}$. Here we investigate the influence of each role. 

For CIFAR100, results are shown in Table~\ref{cifar_role}. The $\mathrm{L}+\mathrm{T}$ setting is similar to DML~\cite{zhang2018deep}. The $\mathrm{L} + \mathrm{A}$ setting is similar to Born again~\cite{furlanello2018born}, where the first generation anchor is a trivially trained model. In contrast, the first generation anchor in  $\mathrm{L} + \mathrm{A} ^*$ is trained with $\mathrm{L} + \mathrm{T}$. For both conditions we report the result after three iterative generations. The result shows that both $\mathrm{A}$ and $\mathrm{T}$ could boost the performance of the target student when introduced individually. However, simply combining these two methods through making the student of $\mathrm{L}+\mathrm{T}$ the first-generation anchor of $\mathrm{L}+\mathrm{A}$ brings minor improvement. Our~\kdname{}, in contrast, further improves the performance of the target student.

For CASIA-Webface, results are shown in Fig.\ref{figure_roles}(a). The Baseline ($\mathrm{L}$) with only task label loss starts in an underfitting state and then grows to an overfitting state. Then, adding only the anchor $\mathrm{L}+\mathrm{A}$ and adding only the teacher $\mathrm{L}+\mathrm{T}$ both bring impressive improvement, illustrating the effectiveness of each role. When including all three roles, further improvement is obtained, clearly illustrating the necessity and effectiveness of the three different roles. We refer readers to Appendix for more ablative experiments. 
\begin{table}[]
  \caption{Effect of each role in triplet distillation. L, T, and A represent the supervision from task label, online teacher, and anchor, respectively. The first-generation anchor in L+A is the model trained with L, while the first-generation anchor in L+T* and L+T+A is trained with L+T. The experiment is conducted on CIFAR100.}
  \label{cifar_role}
\centering
\small
\setlength{\tabcolsep}{1mm}{
\begin{tabular}{c|ccccccc}
\toprule
\multicolumn{1}{c|}{\multirow{2}{*}{Methods}} & resnet56 & wrn-40-2 & wrn-40-2 & resnet32x4 & vgg13 \\
\multicolumn{1}{c|}{}                        & resnet20 & wrn-40-1 & wrn-16-2 & resnet8x4  & vgg8  \\ \midrule
L                                           & 69.29    & 71.63      & 73.47      & 72.92      & 70.10 \\
L+T                                         & 71.22    & 74.73      & 75.41      & 75.36      & 74.58 \\
L+A                                         & 71.70    & 74.06      & 75.18      & 74.35      & 71.63 \\
L+A*                                        & 71.60    & 74.49      & 75.12      & 74.54      & 72.49 \\
L+T+A                                       & 72.34    & 75.96      & 76.94      & 76.82      & 75.35 \\\bottomrule
\end{tabular}}
\vspace{-3mm}
\end{table}

\section{Conclusion}


This work aims to address the problem of the student’s limited ability and the unattainable optimization goal of the large teacher. We propose a novel triplet distillation mechanism,~\kdname{}, to solve the mimicking difficulty problem. Besides teacher and student, we introduce a third model called anchor to make the teacher accurate but easy to mimic. To obtain a high-quality anchor, a curriculum strategy is proposed, which allows the student benefits from accurate but easy-to-mimic hints and obtain good performance, then it can be used as the new anchor for new students. Theoretical analysis in the context of risk minimization decomposition supports the rationality of our method. Furthermore, our TriKD achieves state-of-the-art performance on knowledge distillation and also demonstrates better generalization in tackling the over-fitting issue. In the future, we will explore how we could more efficiently find a proper anchor, and try to extend~\kdname{} to more tasks.

\vfill
\begin{center}
    \noindent{\huge Appendix}
\end{center}

\appendix

\section{Variance and Bias Analysis}
\label{exp:v&b}
In this section, we empirically analyze how~\kdname{} works from a variance-bias perspective. We will show that 1)~\kdname{} reduces the variance of the target student, and 2) a large teacher induces a better-calibrated distribution for the student to mimic, leading to lower bias. We hope the analysis in this section could provide some extra insight. 

According to Proposition 3 in~\cite{menon2020distillation}, for constant $C>0$ and any student network $\mathrm{S}$, the risk in vanilla knowledge distillation could be bounded as:
\begin{equation}
\small
\label{eq:bias_var_sd}
\begin{aligned}
\small
&\mathbb{E}\left[(\tilde{R}(f_\mathrm{S}, D)-R(f_\mathrm{S}))^{2}\right]   \\ 
&\leq \frac{1}{N} \mathbb{V}\left[\mathcal{L}(f_\mathrm{T}(x),f_\mathrm{S}(x)\right] + C \left(\mathbb{E}\left[\left\|f_\mathrm{T}(x)-f_\mathrm{R}(x)\right\|_{2}\right]\right)^{2},
\end{aligned}
\end{equation}
where $\mathbb{E}$ denotes the expectation, $\mathbb{V}$ denotes the variance, $\tilde{R}(\cdot, D)$ is empirical risk on dataset $D$. $\mathcal{L}$ is the distillation loss, typically the KL-Divergence loss. 

In~\kdname{}, there are two types of supervision for the student, \ie{} that from the teacher ($f_{\mathrm{T}}$) and the anchor model ($f_{\mathrm{A}}$), we apply two coefficients ($w_{\mathrm{T}}, w_{\mathrm{A}}$) to combine them, and $w_{\mathrm{T}} + w_{\mathrm{A}} = 1$. Following Eq. (\ref{eq:bias_var_sd}), the variance-bias decomposition of ~\kdname{} is:
\begin{equation}
\label{eq:bias_var_tri}
\begin{aligned}[l]
\small
&\mathbb{E}\left[(\tilde{R}(f_\mathrm{S}, D)-R(f_\mathrm{S}))^{2}\right]   \\ 
&\leq \frac{1}{N} \mathbb{V}\left[\mathcal{L}((w_\mathrm{T}f_\mathrm{T}(x)+w_\mathrm{A}f_\mathrm{A}(x)),f_\mathrm{S}(x)\right] \\
& +  C \left(\mathbb{E}\left[\left\|((w_\mathrm{T}f_\mathrm{T}(x)+w_\mathrm{A}f_\mathrm{A}(x))-f_\mathrm{R}(x)\right\|_{2}\right]\right)^{2}.
\end{aligned}
\end{equation}
This error bound establishes a fundamental variance-bias trade-off when performing distillation. Specifically, they show the fidelity of the distilled risk’s approximation to the expected one mainly depends on two factors: how variable the loss is given a random instance (the variance term), and how well the mimicking target $w_\mathrm{T}f_\mathrm{T}(x)+w_\mathrm{A}f_\mathrm{A}$ approximates the real output $f_\mathrm{R}$ on average (the bias term). Our goal is to analyze how arranging the teacher model $\mathrm{T}$ and the anchor model $\mathrm{A}$ could lower the bound in Eq. (\ref{eq:bias_var_tri}). 

\begin{figure*}[t]
\normalsize
\centering
\includegraphics[width=0.98\linewidth]{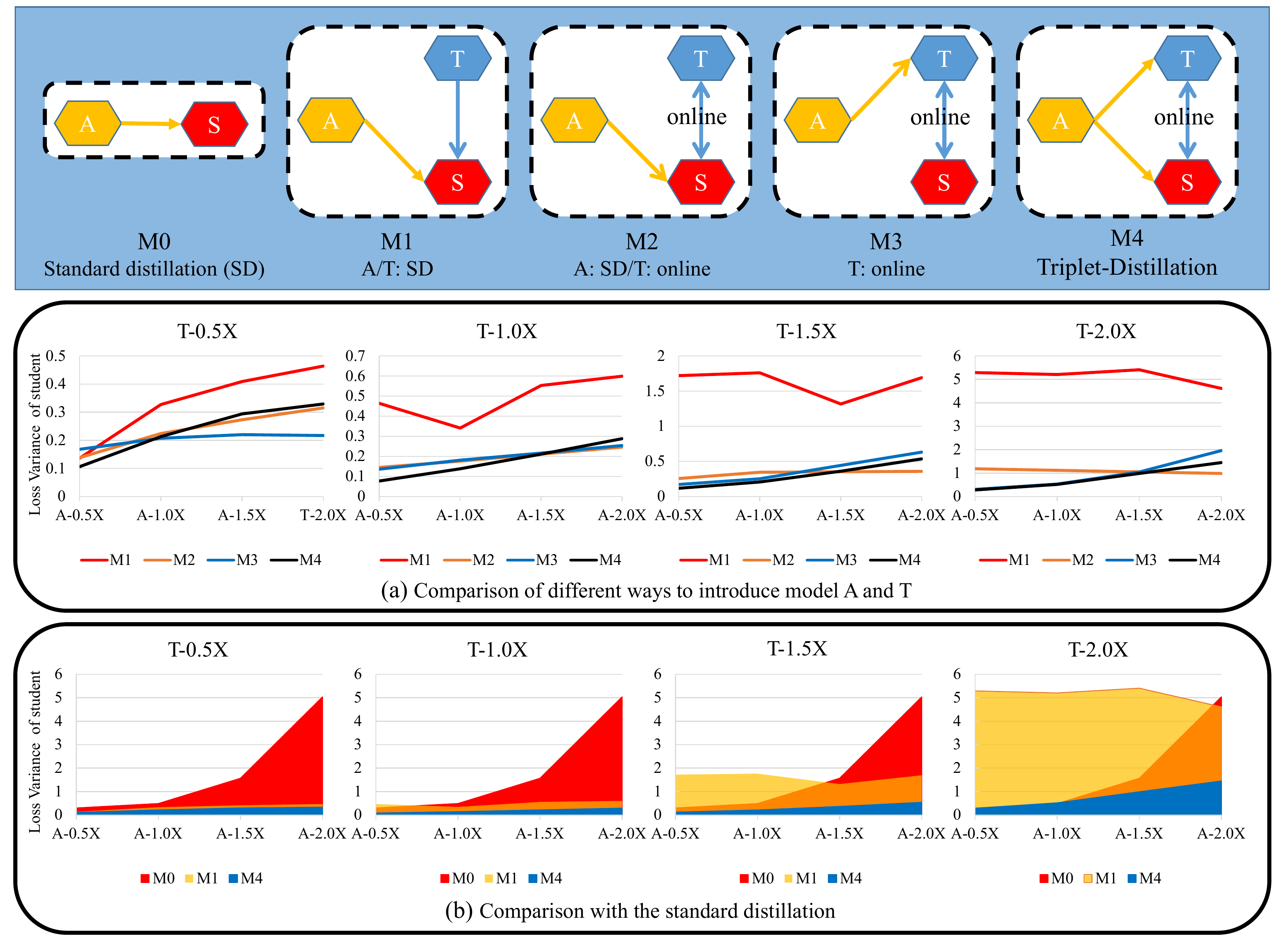}
\vspace{-0.1cm}
\caption{Exploring how to arrange $\mathrm{T}$ and $\mathrm{A}$ to get a lower variance $\mathrm{S}$. (a) and (b) reveal the variance of target model's losses on different conditions. There are basically four valid combinations (\ie{} M1-M4) in terms of two main factors: the way model $\mathrm{S}$ learns from model $\mathrm{T}$ -- standard offline distillation or online mutual learning, and whether model $\mathrm{T}$ learns from model $\mathrm{A}$. Online denotes that two networks study with each other step by step during the training process. \textbf{(a)} illustrates that online mutual learning makes important contribution to decrease the variance, and M4 can gain lower variance when the size of model $\mathrm{A}$ is smaller than model $\mathrm{T}$. \textbf{(b)} demonstrates that M4 can get the lowest variance under all the experimental settings compared with standard distillation (M0 and M1).  Dataset: Webface.}%
\label{fig:theory_part1}
\end{figure*}

For the \textbf{Variance} part, as shown in Fig.\ref{fig:theory_part1}, we conduct experiments to explore how to lower it. There are basically four valid combinations, \ie{} \emph{M0}: $\mathrm{S}$ learns from $\mathrm{A}$ with vanilla distillation, \emph{M1}: $\mathrm{S}$ learns from both $\mathrm{A}$ and $\mathrm{T}$ with vanilla offline distillation, \emph{M2}: $\mathrm{S}$ learns from $\mathrm{A}$ with offline distillation and from $\mathrm{T}$ with online distillation, \emph{M3}: $\mathrm{T}$ learns from $\mathrm{A}$ with vanilla distillation and $\mathrm{S}$ learns from $\mathrm{T}$ with online learning, \emph{M4}: both $\mathrm{S}$ and $\mathrm{T}$ learns from $\mathrm{A}$ with vanilla distillation and $\mathrm{S}$ learns from $\mathrm{T}$ with online learning. Generally, we consider two main factors: the way model $\mathrm{S}$ learns from model $\mathrm{T}$ -- vanilla offline distillation or online mutual distillation, and whether model $\mathrm{T}$ learns from model $\mathrm{A}$. Fig.\ref{fig:theory_part1}(a) reveals that online mutual learning makes important contribution to decrease the variance, and M4, which is used in~\kdname{}, can gain lower variance when the size of model $\mathrm{A}$ is small comparing with $\mathrm{T}$. Furthermore, we compare M4 with vanilla distillation (M0 and M1) as shown in Fig.\ref{fig:theory_part1}(b), M4 can get the lowest variance in all the experiments settings. To sum up, the above experiments show that arranging the anchor $\mathrm{A}$ and the teacher $\mathrm{T}$ as in M4 and making $\mathrm{A}$ small can greatly help reduce the variance. 

\begin{figure*}[t]
\normalsize
\centering
\includegraphics[width=1\linewidth]{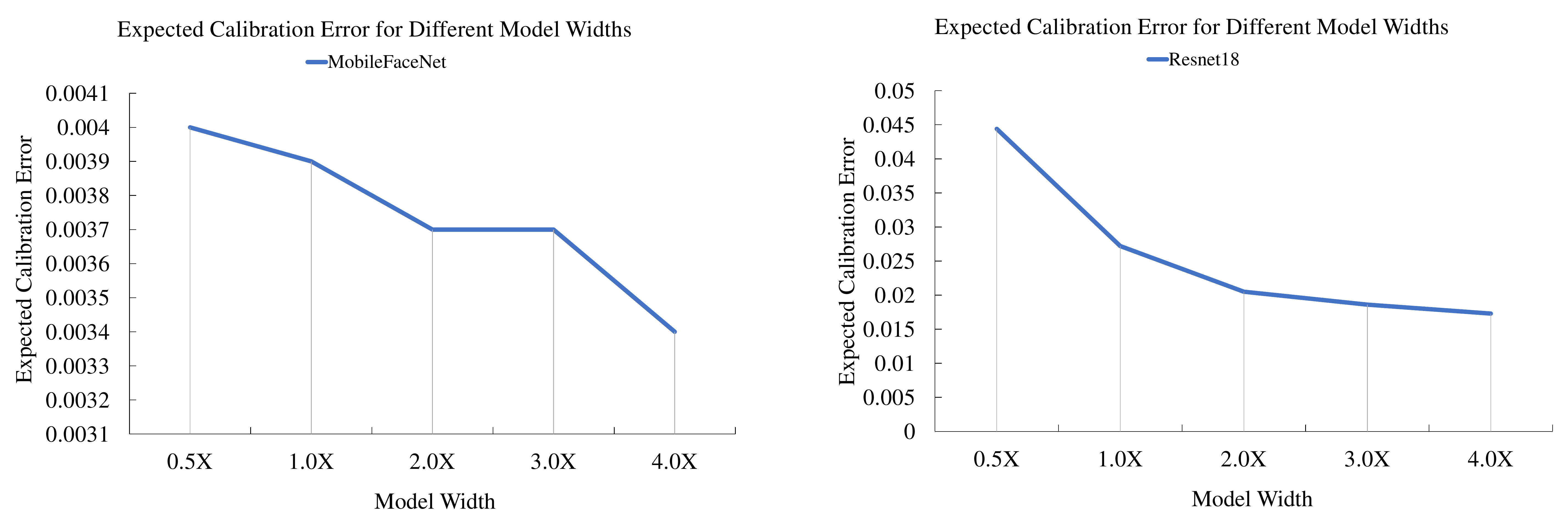}
\vspace{-0.5cm}
\caption{Expected Calibration Error for Different Model Widths. We explore Expected Calibration Error in terms of network width on Face Recognition task (Webface) and Image classification task (CIFAR100), and all the models are trained enough epochs to ensure the model converges sufficiently. The backbones are MobilefaceNet and Resnet18 respectively, we applied various width including $0.5X, 1.0X, 2.0X, 3.0X, 4.0X$.}
\label{fig:theory_part2}
\vspace{-0.3cm}
\end{figure*}

For the \textbf{Bias} part, it follows:
\begin{equation}
\small
\begin{aligned}
 &C \left(\mathbb{E}\left[\left\|((w_\mathrm{T}f_\mathrm{T}(x)+w_\mathrm{A}f_\mathrm{A}(x))-f_\mathrm{R}(x)\right\|_{2}\right]\right)^{2} \\
 &\leq C \left(\mathbb{E}\left[w_\mathrm{T}\|(f_\mathrm{T}(x)-f_\mathrm{R}(x)\|_2+w_\mathrm{A}\|f_\mathrm{A}(x)-f_\mathrm{R}(x)\|_{2}\right]\right)^{2}.
\end{aligned}
\end{equation}
The second line is obtained based on triangular inequality. Minimizing this term means that we should make the introduced teacher model $\mathrm{T}$ as well as the anchor model $\mathrm{A}$ approximate the Bayes class-probability distribution $f_\mathrm{R}$ better. In detail, it means the expected calibration error (ECE)~\cite{ece} of the two models should be small. In~\cite{guo2017calibration}, the authors analysed the calibration measured by ECE in terms of different aspects, such as network depth, width, Batch Normalization and weight decay. The experiments in~\cite{guo2017calibration} showed that increasing width of a network will make the ECE first rise and then fall. To make it clearer, we conduct this experiment again in terms of the effect of network width on face recognition task (Webface) and image classification task (CIFAR100), and all the models are trained enough epochs to ensure the model converges sufficiently. The backbones are MobilefaceNet and Resnet18 respectively, we applied various width including $0.5X, 1.0X, 2.0X, 3.0X, 4.0X$. As shown in Fig.\ref{fig:theory_part2}, we observe that increasing the network width positively affect model calibration. As a result, we can minimize the bias term through making the model $\mathrm{T}$ wider. The anchor $\mathrm{A}$, however, faces a variance-bias trade-off: as shown in the variance part, small anchor tend to benefit lowering the variance, but it could degrade the bias, and vice versa. In this paper, we keep the anchor $\mathrm{A}$ small (the same size as the student) in favor of low variance, and we leave further exploration of the trade-off to future work. Combining the above two parts, we can introduce a large model $\mathrm{T}$ to M4, and keep the anchor $\mathrm{A}$ small, which forms our proposed~\kdname{}.


\section{Experimental Details}
\noindent\textbf{CIFAR100}~\cite{krizhevsky2009learning} dataset consists of 60K images from 100 categories with size of $32\times32$. In the standard protocol, 50k images are used for training and 10K for testing. We choose CIFAR-style resnet~\cite{he2016deep}, wide-resnet~\cite{wrn} and vgg~\cite{vgg} as model architecture. We train all the models for 240 epochs. The initial learning rate is 0.1 and is decayed by a factor of 10 at 150, 180, and 210 epochs, respectively. We run experiments on one Tesla-V100 GPU with a batch size of 128. An SGD optimizer with 0.0005 weight decay and 0.9 momentum is adopted. For all the experiments, we set $w_1 = w_2 = w_3 = w_4 = w_5 = w_6=1$ at the beginning. After epoch 150, where the learning rate decays for the first time, we decrease $w_1$ to $0.1$ and increase $w_2$ to $10$. For all experiments except vgg, the temperature $\tau$ is set to 1 for $\mathcal{L}_{KL}$; for vgg, we set it to 4.

\noindent\textbf{ImageNet}~\cite{deng2009imagenet} consists of 1.28 million training images and 50k validation images from 1000 categories. Following the mainstream settings, all methods are trained on the entire training set and evaluated on the single-crop validation set. The input image resolution is $224\times224$ for both training and evaluation. We use resnet34 as teacher and resnet18 as student. We train all the models for 100 epochs. The initial learning rate is 0.1 and is decayed by a factor of 10 at 30, 60, and 90 epochs, respectively. We run experiments on one Tesla-V100 GPU with a batch size of 256. An SGD optimizer with a 0.0001 weight decay and 0.9 momentum is adopted. Due to limited resources, we simply set $w_1 = w_2 = w_3 = w_4 = w_5 = w_6=1$, and $\tau=1$.

\noindent\textbf{CASIA-WebFace}~\cite{yi2014learning} consists of 494,414 face images from 10,575 identities. Besides the full training set, two subsets of 50k and 150k images are randomly selected for efficient training. \textbf{MegaFace}~\cite{kemelmacher2016megaface} dataset is used for testing, which contains 1M images of 60k identities as the gallery set and 100k images of 530 identities from FaceScrub as the probe set. For better stability of training, Arcface loss~\cite{deng2019arcface} used in MobileFaceNet is replaced with AM-Softmax loss~\cite{wang2018additive} in our experiments. Following the work of AM-Softmax loss, the faces are aligned and cropped out with size of $112\times96$.
For optimization, SGD with momentum 0.9 is used and the batch size is 256. All the models are trained with 40k iterations. The learning rate starts from 0.1 and linearly reduces to 0. The setting of weight decay keeps the same as~\cite{chen2018mobilefacenets}.

\section{More experiments}
\subsection{Comparing with TAKD}
\label{sec_takd_compare}
Large models tend to generalize better. However, existing studies~\cite{teacherassistant, StudentCustomized, ESKD} have shown that in knowledge distillation, the performance of the student would indeed deteriorate when the capacity of the teacher increases. To boost the performance of the student when the capacity gap between the teacher and the student is large, TAKD~\cite{teacherassistant} proposed to bridge the gap by introducing intermediate-sized models named teacher assistant. Both TAKD and our~\kdname{} attempt to reduce the difficulty for the student to mimic the teacher. However, TAKD treats learning difficulty as an inherent property of teacher model capacity,~\ie{} larger teachers are inherently harder, and smaller teachers are easier. In contrast, we believe that a given network architecture with fixed capacity should be able to fit both hard and easy functions, and we could make a large teacher still easy to mimic by deliberately making the function it expresses easy; the reason why large teacher usually fails in existing distillation frameworks is that the teacher would spontaneously learn to express sophisticated functions when trained without constraint. This is easy to understand when considering the teacher model's function identity: with larger capacity, the larger teacher should be able to easily fit the same function as a smaller teacher does, and thus in distillation a student supervised by a larger teacher should at least perform no worse than supervised by a smaller one. Here we also provide an experiment to compare our~\kdname{} with TAKD. The experiment is conducted on CIFAR100. For fair comparision, following TAKD, we use resnet8 as the student and resnet110 as the teacher, and we use stochastic gradient descent with Nesterov momentum of 0.9 and learning rate of 0.1 for 150 epochs. we decrease learning rate to 0.01 on epoch 80 and 0.001 on epoch 120. Weight decay is set to 0.0001. The result is shown in Table~\ref{suptable:takd}. Its shows that our~\kdname{} consistently outperforms TAKD with different teacher assistant size.

We further emphasize that our proposed~\kdname{} is a general knowledge distillation method rather than specially designed for situations where the capacity gap between the teacher and the student is large, like~\cite{teacherassistant, ESKD, StudentCustomized}. The mimicking difficulty is a ubiquitous problem in knowledge distillation rather than exclusive to teacher-student pairs with extremely large capacity gap. Experiments also show that this method could greatly benefit the student even though the teacher is relatively small.
\begin{table}[h]
  \small
  \caption{Compare~\kdname{} with KD~\cite{hinton2015distilling} and TAKD~\cite{teacherassistant}. Dataset: CIFAR100. Student=resnet8, Teacher=resnet110. The results of KD and TAKD are quoted from the original TAKD paper.}
  \label{suptable:takd}
  \centering
  \begin{tabular}{c|cccc|c}
    \toprule
    \multicolumn{1}{c}{\multirow{2}{*}{KD}}&\multicolumn{4}{|c|}{TAKD} 
    &\multicolumn{1}{c}{\multirow{2}{*}{~\kdname{}}} \\
    \cmidrule{2-5} 
    &TA=56 & TA=32 & TA=20 & TA=14 & \\
     \midrule
    61.41 & 61.47 & 61.55 & 61.82 & 61.50 & 62.79\\
    \bottomrule
  \end{tabular}
  \vspace{-0.2cm}
\end{table}

\begin{table*}[h]
  \normalsize
  \caption{Additional results of~\kdname{} w.r.t. different network architectures. Teacher is two times as wide as the student.}
  \label{suptable:additional}
  \centering
  \begin{tabular*}{\textwidth}{c@{\extracolsep{\fill}}|cccc|cccc}
    \toprule
    \multicolumn{1}{c|}{} 
    &\multicolumn{4}{c|}{CIFAR100} 
    &\multicolumn{4}{c}{ImageNet} \\
    \midrule
    Backbone     & MobileV2   & ResNet18 &  ResNet34 & ResNet50 & MobileV1 & MobileV2 & ShuffleV2 & ResNet18  \\
    (Madds)    & (90M) & (555M) & (1.16G) & (1.30G) & (569M) & (300M) & (147M) & (2.34G) \\
    \midrule
    Baseline & 72.0 & 77.4 & 77.9& 77.4 & 71.8 & 72.6 & 68.9 & 71.0\\
    \kdname{}(Ours) & 75.1 & 79.3 & 80.3 & 79.4 & 74.2 & 73.8 & 70.6 & 72.7 \\
    \bottomrule 
  \end{tabular*}
  \vspace{-0.1cm}
\end{table*}

\begin{table*}[]
  \normalsize
  \caption{Performance of target student ($\mathrm{S}$) w.r.t. different model size of online teacher ($\mathrm{T}$), e.g. 0.5X/1.0X/2.0X. Baseline means trained with only hard label. Dataset: WebFace. Network: MobileFaceNet.}
  \label{top_student_size}
  \centering
  \begin{tabular}{c|cccc|ccc|c}
    \toprule
    \multicolumn{1}{c|}{\multirow{2}*{Student}} 
    &\multicolumn{4}{c|}{Rank-1 identification rate of $\mathrm{S}$(\%)} 
    &\multicolumn{3}{c|}{Rank-1 identification rate of $\mathrm{T}$(\%)} 
    &\multicolumn{1}{c}{\multirow{2}*{Madds}} \\
    \cmidrule{2-8}
    & Baseline
    &\multicolumn{1}{c}{\ \  $\mathrm{T}$=0.5X \ \ } & {\ \ $\mathrm{T}$=1.0X \ \ } & {\ \ $\mathrm{T}$=2.0X\ \ } & {\ \ $\mathrm{T}$=0.5X\ \ } & {\ \ $\mathrm{T}$=1.0X\ \ } & {\ \ $\mathrm{T}$=2.0X\ \ }
    &\multicolumn{1}{c}{} \\
    \midrule
    0.50X&64.0&63.2&67.6&69.0&65.0&73.5&75.8&50M \\
    0.75X&68.3&68.6&74.0&75.7&68.0&77.1&79.7&109M \\
    1.00X&69.7&71.7&77.4&79.3&68.8&77.8&80.7&189M \\
    1.25X&69.4&73.2&79.5&81.5&68.7&78.4&81.6&292M \\
    1.50X&68.3&74.6&81.0&82.4&69.6&78.5&81.5&487M \\
    \bottomrule
  \end{tabular}
\end{table*}

\begin{table*}[]
  \normalsize
  \centering

  \caption{Best accuracy(\%) achieved by student after each generation. Except generation 0, where we use vanilla online distillation to train an initial anchor, for all generations we use the last-generation student as the anchor, and use randomly initialized student and teacher to form the triplet relationship. The experiment is conducted on CIFAR100.}
  \label{generation_count_table}
\begin{tabular}{c|ccccccc}
\toprule
\multirow{2}{*}{Generations} & resnet56 & resnet110 & resnet110 & wrn-40-2 & wrn-40-2 & resnet32x4 & vgg13 \\
                             & resnet20 & resnet20  & resnet32  & wrn-40-1 & wrn-16-2 & resnet8x4  & vgg8  \\
\midrule
0                            & 71.22    & 71.47     & 73.52     & 74.73      & 75.41      & 75.36      & 74.58       \\
1                            & 71.76    & 71.82     & 73.99     & 75.35      & 76.94      & 76.27      & 75.35       \\
2                            & 72.34    & 72.24     & 74.31     & 75.87      & 76.94      & 76.82      & 75.35       \\
3                            & 72.34    & 72.55     & 74.31     & 75.96      & 76.94      & 76.82      & 75.35       \\
4                            & 72.34    & 72.55     & 74.31     & 75.96      & 76.94      & 76.82      & 75.35       \\ \bottomrule
\end{tabular}
\end{table*}

\subsection{Additional Results on Image Classification}
We provide some additional results with more architectures on image classification. For the experiments in this section, we set the teacher to be 2 times as wide as the student. For experiments on ImageNet, all methods are trained for 120 epochs. For the hyper-parameters, SGD with momentum 0.9 is used for optimization and the batch size is 256. The learning rate starts from 0.1 and linearly reduces to 0. The weight decay set as $5e-4$ for ShuffleNet V2, $1e-4$ for ResNet18. For experiments on CIFAR100, all models are trained for 200 epochs. As for the hyper-parameters, SGD with momentum 0.9 is used for optimization and the batch size is 128. The learning rate starts from 0.1 and is multiplied by 0.1 at 60, 120 and 180 epochs. The weight decay is set as $5e-4$. Table~\ref{suptable:additional} shows the result.

\subsection{Impact of Teacher Size}
The teacher, a large network with high fitting ability, represents the potential upper limit of student's performance. Without losing flexibility, it can be set with any desired model size no less than the target model size. Table~\ref{top_student_size} shows the results of our~\kdname{} with the teacher in different model size, \ie{} $0.5\times$, $1.0\times$, $2.0\times$ of the base network size. The experiment is conducted on face recognition and the network architecture is MobileFaceNet. As can be seen, our learning mechanism is stable w.r.t. different size of the teacher models, which can flexibly adapt to different training resources and better meet the trade-off between computational cost and performance. More specifically, larger teacher $\mathrm{T}$ induce better model $\mathrm{S}$, which is consistent with our motivation and demonstrates that larger model $\mathrm{T}$ has an edge in exploring generalizable solutions. 

\begin{figure}[t]
\normalsize
\centering
\includegraphics[width=1\linewidth]{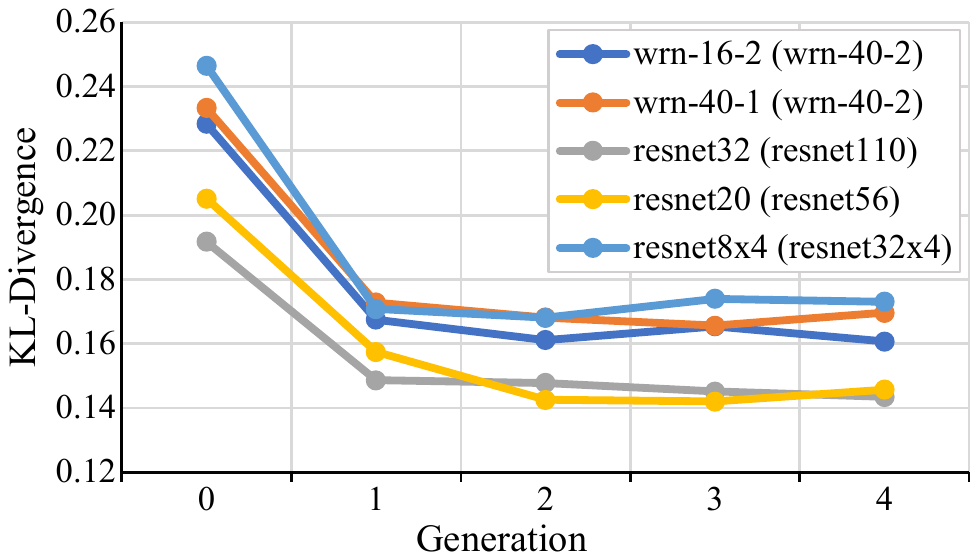}
\caption{Teacher-student behavior similarity w.r.t. generations. Generation 0 is vanilla online knowledge distillation without anchor. The networks are trained on the training set of CIFAR100, and KL-Divergence is measured on the test set of CIFAR100. Legend format: student (teacher).}
\label{fig:similarity}
\end{figure}

\subsection{Iterate for different number of generations}
As mentioned in~\ref{iterationsec}, we adopt a curriculum strategy to obtain an appropriate anchor model for~\kdname{}. Here we investigate how many generations are needed for this process. The experiment is conducted on CIFAR100. Table~\ref{generation_count_table} shows the results. Generation 0, as mentioned in~\ref{iterationsec}, is a plain online distillation process without using an anchor. The result shows that it generally takes 1 to 2 generations (generation 0 not included) for the process to converge, and at that time the student generally reaches a good performance. We empirically find that the first and the second generations are the most likely to bring in improvement, and the following generations tend to bring in less, if any. Specifically, we attribute the improvement in the first and later generations to different mechanisms. The first generation's improvement is due to the introduction of the triplet relationship, and the later generations improves the student through using more accurate anchor; the former is qualitative, and the latter is majorly quantitative. As shown in Fig.\ref{fig:similarity}, from a teacher-student behavior similarity perspective, the KL-divergence between the teacher and the student drops dramatically after generation 1, but then drops slowly in the following generations. It means that it is the triplet relationship, rather than the curriculum process, that makes the mimicking easier. On the other hand, from the variance-bias perspective (see~\ref{exp:v&b}), the curriculum learning can be identified as a means to gradually decrease the bias of the anchor.

{
\small
\bibliography{example_paper}

\begin{thebibliography}{39}
\providecommand{\natexlab}[1]{#1}
\providecommand{\url}[1]{\texttt{#1}}
\expandafter\ifx\csname urlstyle\endcsname\relax
  \providecommand{\doi}[1]{doi: #1}\else
  \providecommand{\doi}{doi: \begingroup \urlstyle{rm}\Url}\fi

\bibitem[Ahn et~al.(2019)Ahn, Hu, Damianou, Lawrence, and Dai]{vid}
Ahn, S., Hu, S.~X., Damianou, A., Lawrence, N.~D., and Dai, Z.
\newblock Variational information distillation for knowledge transfer.
\newblock In \emph{IEEE/CVF Conference on Computer Vision and Pattern
  Recognition (CVPR)}, pp.\  9163--9171, 2019.

\bibitem[Chen et~al.(2020)Chen, Mei, Wang, Feng, and Chen]{okddip}
Chen, D., Mei, J.-P., Wang, C., Feng, Y., and Chen, C.
\newblock Online knowledge distillation with diverse peers.
\newblock In \emph{AAAI Conference on Artificial Intelligence}, pp.\
  3430--3437, 2020.

\bibitem[Chen et~al.(2021)Chen, Liu, Zhao, and Jia]{review}
Chen, P., Liu, S., Zhao, H., and Jia, J.
\newblock Distilling knowledge via knowledge review.
\newblock In \emph{IEEE/CVF Conference on Computer Vision and Pattern
  Recognition (CVPR)}, pp.\  5008--5017, 2021.

\bibitem[Chen et~al.(2018)Chen, Liu, Gao, and Han]{chen2018mobilefacenets}
Chen, S., Liu, Y., Gao, X., and Han, Z.
\newblock Mobilefacenets: Efficient cnns for accurate real-time face
  verification on mobile devices.
\newblock In \emph{Chinese Conference on Biometric Recognition (CCBR)}, pp.\
  428--438, 2018.

\bibitem[Cho \& Hariharan(2019)Cho and Hariharan]{ESKD}
Cho, J.~H. and Hariharan, B.
\newblock On the efficacy of knowledge distillation.
\newblock In \emph{IEEE/CVF International Conference on Computer Vision
  (ICCV)}, pp.\  4794--4802, 2019.

\bibitem[Coates et~al.(2011)Coates, Ng, and Lee]{stl10}
Coates, A., Ng, A., and Lee, H.
\newblock An analysis of single-layer networks in unsupervised feature
  learning.
\newblock In \emph{International Conference on Artificial Intelligence and
  Statistics}, pp.\  215--223, 2011.

\bibitem[Deng et~al.(2009)Deng, Dong, Socher, Li, Li, and
  Fei-Fei]{deng2009imagenet}
Deng, J., Dong, W., Socher, R., Li, L.-J., Li, K., and Fei-Fei, L.
\newblock Imagenet: A large-scale hierarchical image database.
\newblock In \emph{IEEE Conference on Computer Vision and Pattern Recognition
  (CVPR)}, pp.\  248--255, 2009.

\bibitem[Deng et~al.(2019)Deng, Guo, Xue, and Zafeiriou]{deng2019arcface}
Deng, J., Guo, J., Xue, N., and Zafeiriou, S.
\newblock Arcface: Additive angular margin loss for deep face recognition.
\newblock In \emph{IEEE/CVF Conference on Computer Vision and Pattern
  Recognition (CVPR)}, pp.\  4690--4699, 2019.

\bibitem[Ding et~al.(2019)Ding, Wu, Sun, Guo, and Xia]{ding2019adaptive}
Ding, Q., Wu, S., Sun, H., Guo, J., and Xia, S.-T.
\newblock Adaptive regularization of labels.
\newblock \emph{arXiv preprint arXiv:1908.05474}, 2019.

\bibitem[Furlanello et~al.(2018)Furlanello, Lipton, Tschannen, Itti, and
  Anandkumar]{furlanello2018born}
Furlanello, T., Lipton, Z.~C., Tschannen, M., Itti, L., and Anandkumar, A.
\newblock Born again neural networks.
\newblock In \emph{International Conference on Machine Learning (ICML)}, pp.\
  1607--1616, 2018.

\bibitem[Guo et~al.(2017)Guo, Pleiss, Sun, and Weinberger]{guo2017calibration}
Guo, C., Pleiss, G., Sun, Y., and Weinberger, K.~Q.
\newblock On calibration of modern neural networks.
\newblock In \emph{International Conference on Machine Learning (ICML)}, pp.\
  1321--1330, 2017.

\bibitem[Guo et~al.(2020)Guo, Wang, Wu, Yu, Liang, Hu, and Luo]{KDCL}
Guo, Q., Wang, X., Wu, Y., Yu, Z., Liang, D., Hu, X., and Luo, P.
\newblock Online knowledge distillation via collaborative learning.
\newblock In \emph{IEEE/CVF Conference on Computer Vision and Pattern
  Recognition (CVPR)}, pp.\  11020--11029, 2020.

\bibitem[He et~al.(2016)He, Zhang, Ren, and Sun]{he2016deep}
He, K., Zhang, X., Ren, S., and Sun, J.
\newblock Deep residual learning for image recognition.
\newblock In \emph{IEEE Conference on Computer Vision and Pattern Recognition
  (CVPR)}, pp.\  770--778, 2016.

\bibitem[Hinton et~al.(2015)Hinton, Vinyals, and Dean]{hinton2015distilling}
Hinton, G., Vinyals, O., and Dean, J.
\newblock Distilling the knowledge in a neural network.
\newblock \emph{Advances in Neural Information Processing Systems (NIPS)},
  2015.

\bibitem[Jin et~al.(2019)Jin, Peng, Wu, Liu, Liu, Liang, Yan, and
  Hu]{jin2019knowledge}
Jin, X., Peng, B., Wu, Y., Liu, Y., Liu, J., Liang, D., Yan, J., and Hu, X.
\newblock Knowledge distillation via route constrained optimization.
\newblock In \emph{IEEE/CVF International Conference on Computer Vision
  (ICCV)}, pp.\  1345--1354, 2019.

\bibitem[Kemelmacher-Shlizerman et~al.(2016)Kemelmacher-Shlizerman, Seitz,
  Miller, and Brossard]{kemelmacher2016megaface}
Kemelmacher-Shlizerman, I., Seitz, S.~M., Miller, D., and Brossard, E.
\newblock The megaface benchmark: 1 million faces for recognition at scale.
\newblock In \emph{IEEE Conference on Computer Vision and Pattern Recognition
  (CVPR)}, pp.\  4873--4882, 2016.

\bibitem[Kim et~al.(2018)Kim, Park, and Kwak]{FT}
Kim, J., Park, S., and Kwak, N.
\newblock Paraphrasing complex network: Network compression via factor
  transfer.
\newblock \emph{Advances in Neural Information Processing Systems (NIPS)}, pp.\
   2765--2774, 2018.

\bibitem[Krizhevsky et~al.(2009)Krizhevsky, Hinton,
  et~al.]{krizhevsky2009learning}
Krizhevsky, A., Hinton, G., et~al.
\newblock Learning multiple layers of features from tiny images.
\newblock 2009.

\bibitem[Lopez-Paz et~al.(2015)Lopez-Paz, Bottou, Sch{\"o}lkopf, and
  Vapnik]{unifying_vapnik}
Lopez-Paz, D., Bottou, L., Sch{\"o}lkopf, B., and Vapnik, V.
\newblock Unifying distillation and privileged information.
\newblock \emph{arXiv preprint arXiv:1511.03643}, 2015.

\bibitem[Menon et~al.(2020)Menon, Rawat, Reddi, Kim, and
  Kumar]{menon2020distillation}
Menon, A.~K., Rawat, A.~S., Reddi, S.~J., Kim, S., and Kumar, S.
\newblock Why distillation helps: a statistical perspective.
\newblock \emph{arXiv preprint arXiv:2005.10419}, 2020.

\bibitem[Mirzadeh et~al.(2020)Mirzadeh, Farajtabar, Li, Levine, Matsukawa, and
  Ghasemzadeh]{teacherassistant}
Mirzadeh, S.~I., Farajtabar, M., Li, A., Levine, N., Matsukawa, A., and
  Ghasemzadeh, H.
\newblock Improved knowledge distillation via teacher assistant.
\newblock In \emph{AAAI Conference on Artificial Intelligence}, pp.\
  5191--5198, 2020.

\bibitem[Naeini et~al.(2015)Naeini, Cooper, and Hauskrecht]{ece}
Naeini, M.~P., Cooper, G., and Hauskrecht, M.
\newblock Obtaining well calibrated probabilities using bayesian binning.
\newblock In \emph{AAAI Conference on Artificial Intelligence}, pp.\
  2901--2907, 2015.

\bibitem[Netzer et~al.(2011)Netzer, Wang, Coates, Bissacco, Wu, and Ng]{SVHN}
Netzer, Y., Wang, T., Coates, A., Bissacco, A., Wu, B., and Ng, A.~Y.
\newblock Reading digits in natural images with unsupervised feature learning.
\newblock In \emph{NIPS Workshop on Deep Learning and Unsupervised Feature
  Learning}, 2011.

\bibitem[Romero et~al.(2015)Romero, Ballas, Kahou, Chassang, Gatta, and
  Bengio]{romero2014fitnets}
Romero, A., Ballas, N., Kahou, S.~E., Chassang, A., Gatta, C., and Bengio, Y.
\newblock Fitnets: Hints for thin deep nets.
\newblock In \emph{International Conference on Learning Representations
  (ICLR)}, 2015.

\bibitem[Simonyan \& Zisserman(2014)Simonyan and Zisserman]{vgg}
Simonyan, K. and Zisserman, A.
\newblock Very deep convolutional networks for large-scale image recognition.
\newblock \emph{arXiv preprint arXiv:1409.1556}, 2014.

\bibitem[Stanton et~al.(2021)Stanton, Izmailov, Kirichenko, Alemi, and
  Wilson]{reallywork}
Stanton, S., Izmailov, P., Kirichenko, P., Alemi, A.~A., and Wilson, A.~G.
\newblock Does knowledge distillation really work?
\newblock \emph{Advances in Neural Information Processing Systems (NeurIPS)},
  pp.\  6906--6919, 2021.

\bibitem[Tian et~al.(2020)Tian, Krishnan, and Isola]{tian2019contrastive}
Tian, Y., Krishnan, D., and Isola, P.
\newblock Contrastive representation distillation.
\newblock In \emph{International Conference on Learning Representations
  (ICLR)}, 2020.

\bibitem[Wang et~al.(2018)Wang, Cheng, Liu, and Liu]{wang2018additive}
Wang, F., Cheng, J., Liu, W., and Liu, H.
\newblock Additive margin softmax for face verification.
\newblock \emph{IEEE Signal Processing Letters (SPL)}, pp.\  926--930, 2018.

\bibitem[Wen et~al.(2019)Wen, Lai, and Qian]{wen2019preparing}
Wen, T., Lai, S., and Qian, X.
\newblock Preparing lessons: Improve knowledge distillation with better
  supervision.
\newblock \emph{arXiv preprint arXiv:1911.07471}, 2019.

\bibitem[Xu et~al.(2020)Xu, Liu, Li, and Loy]{SSKD}
Xu, G., Liu, Z., Li, X., and Loy, C.~C.
\newblock Knowledge distillation meets self-supervision.
\newblock In \emph{European Conference on Computer Vision (ECCV)}, pp.\
  588--604, 2020.

\bibitem[Yao \& Sun(2020)Yao and Sun]{DCM}
Yao, A. and Sun, D.
\newblock Knowledge transfer via dense cross-layer mutual-distillation.
\newblock In \emph{European Conference on Computer Vision (ECCV)}, pp.\
  294--311, 2020.

\bibitem[Yi et~al.(2014)Yi, Lei, Liao, and Li]{yi2014learning}
Yi, D., Lei, Z., Liao, S., and Li, S.~Z.
\newblock Learning face representation from scratch.
\newblock \emph{arXiv preprint arXiv:1411.7923}, 2014.

\bibitem[Zagoruyko \& Komodakis(2016)Zagoruyko and Komodakis]{wrn}
Zagoruyko, S. and Komodakis, N.
\newblock Wide residual networks.
\newblock \emph{arXiv preprint arXiv:1605.07146}, 2016.

\bibitem[Zagoruyko \& Komodakis(2017)Zagoruyko and
  Komodakis]{zagoruyko2016paying}
Zagoruyko, S. and Komodakis, N.
\newblock Paying more attention to attention: Improving the performance of
  convolutional neural networks via attention transfer.
\newblock In \emph{International Conference on Learning Representations
  (ICLR)}, 2017.

\bibitem[Zhang et~al.(2019)Zhang, Song, Gao, Chen, Bao, and Ma]{zhang2019your}
Zhang, L., Song, J., Gao, A., Chen, J., Bao, C., and Ma, K.
\newblock Be your own teacher: Improve the performance of convolutional neural
  networks via self distillation.
\newblock In \emph{IEEE/CVF International Conference on Computer Vision
  (ICCV)}, pp.\  3713--3722, 2019.

\bibitem[Zhang et~al.(2020)Zhang, Lu, Gong, Luo, and Liu]{AMLN}
Zhang, X., Lu, S., Gong, H., Luo, Z., and Liu, M.
\newblock Amln: adversarial-based mutual learning network for online knowledge
  distillation.
\newblock In \emph{European Conference on Computer Vision (ECCV)}, pp.\
  158--173, 2020.

\bibitem[Zhang et~al.(2018)Zhang, Xiang, Hospedales, and Lu]{zhang2018deep}
Zhang, Y., Xiang, T., Hospedales, T.~M., and Lu, H.
\newblock Deep mutual learning.
\newblock In \emph{IEEE/CVF Conference on Computer Vision and Pattern
  Recognition (CVPR)}, pp.\  4320--4328, 2018.

\bibitem[Zhao et~al.(2022)Zhao, Cui, Song, Qiu, and Liang]{DKD}
Zhao, B., Cui, Q., Song, R., Qiu, Y., and Liang, J.
\newblock Decoupled knowledge distillation.
\newblock In \emph{IEEE/CVF Conference on Computer Vision and Pattern
  Recognition (CVPR)}, pp.\  11953--11962, 2022.

\bibitem[Zhu \& Wang(2021)Zhu and Wang]{StudentCustomized}
Zhu, Y. and Wang, Y.
\newblock Student customized knowledge distillation: Bridging the gap between
  student and teacher.
\newblock In \emph{IEEE/CVF International Conference on Computer Vision
  (CVPR)}, pp.\  5057--5066, 2021.

\end{thebibliography}
\bibliographystyle{icml2023}
}
\end{document}